\documentclass[10pt,twocolumn,letterpaper]{article}

\usepackage[pagenumbers]{cvpr}

\definecolor{cvprblue}{rgb}{0.21,0.49,0.74}
\usepackage[pagebackref,breaklinks,colorlinks,allcolors=cvprblue]{hyperref}
\hypersetup{
  pdftitle={Measuring and Predicting Image-Token Repair Boundaries in Vision-Language Models},
  pdfauthor={Vishal Bhutani, Prasang Gupta, Sumanth Chundru, Kevin Paul}
}

\def\confName{CVPR}
\def\confYear{2027}
\graphicspath{{figures/}}

\title{Measuring and Predicting Image-Token Repair Boundaries in Vision-Language Models}

\author{
Vishal Bhutani \quad Prasang Gupta \quad Sumanth Chundru \quad Kevin Paul\\
PricewaterhouseCoopers (PwC)\\
{\tt\small
\{vishal.bhutani, prasang.gupta, sumanth.s.chundru, kevin.x.paul\}@pwc.com}
}

\begin{document}
\maketitle

\begin{abstract}
Degraded images make vision-language models answer wrongly. Using synthetic paired blur, noise, and low light, we trace which clean image states can still repair those answers from the vision encoder through the language model. In four encoder models, 12--23 ranked tokens recover half of the full-patch effect, whereas random orders require four to fifteen times as many; encoder-free Gemma needs 22.2 versus 53.1. Full-block repair expires at a model-specific language-model crossing. The matched reverse-direction intervention is algebraically identical to text-token restoration, so its gap from image restoration measures a redundancy band rather than an independent predictor. A clean-image gray-ablation crossing independently predicts held-out repair with 1.53-layer error, against 3.65 for shared depth. Gemma MMBench has the only large attention-to-repair point gap; Gemma GQA does not. Restormer provides a non-oracle consistency check on two models.
\end{abstract}

\section{Introduction}
\label{sec:intro}

\begin{figure}[t]
  \centering
  \includegraphics[width=\linewidth]{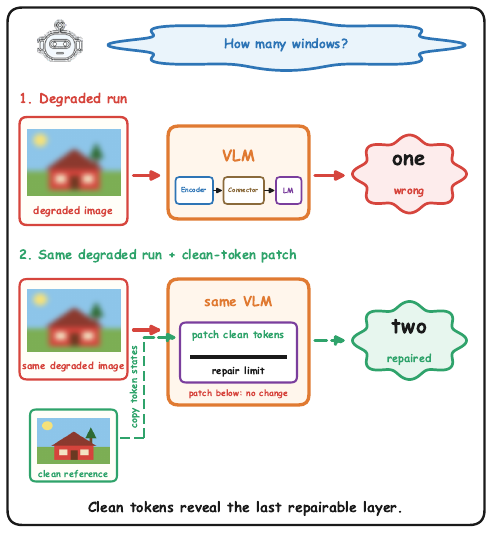}
  \caption{The model sees only the degraded image and answers one instead of two. The clean image is a reference, not a second model input: its token states are pasted into the degraded run. The answer returns when that patch is made before the repair boundary; the same patch at a later layer changes nothing.}
  \label{fig:hook}
\end{figure}

A blurred, noisy, or poorly lit photograph makes a vision-language model answer a question it would have got right on the clean image~\cite{hendrycks2019robustness,qiu2025benchmarking}. Accuracy says how far the answer falls. We take the same question, keep the degraded photo, and put selected image tokens back to the values they had on the clean photo. If the answer returns, the damage can still be repaired at that spot. Repair means restoring those tokens inside the model.

The photo first becomes a few hundred image tokens in the vision encoder. In a five-model stage-map experiment, we rank those tokens by how strongly each one pushes the model toward the answer it gave on the clean image, then put the clean values back in that order. Restoring 12--23 top-ranked encoder tokens recovers half of what restoring the full token set recovers; ten random orders need four to fifteen times as many. Gemma's first repairable language-model locus needs 22.2 ranked tokens versus 53.1 random. The connector then merges neighboring tokens into a shorter sequence. After that merge, a handful of attributed tokens reaches the same half-recovery level on the models that merge.

We then walk the restoration through the language model. The \emph{repair limit} is the layer where full image-token restoration falls below half of its connector-input ceiling. A complementary text-token patch rises over the same layers; because the two position sets partition the prompt, we use that curve as a consistency check. Attention and repair agree within the predeclared margin for LLaVA and are inconclusive for Qwen2-VL. Gemma MMBench has a 4.14-layer point gap, but one of three condition intervals crosses the margin; Gemma GQA has no separation.

We test this window across seven decoder-only architectures under conditions calibrated toward a 15-point accuracy drop. On the normalized primary populations, blur, noise, and low light span at most 1.44 layers on MMBench and 1.10 on GQA. Raw-threshold population variants reach 1.52. The normalized 42-condition MMBench/GQA slope analysis detects no damage trend inside this selected synthetic band. Six models also have similar point estimates on MMBench and GQA. Encoder-free Gemma remains task-dependent: its dense MMBench estimate is 4.02 layers later than on GQA; CV-Bench has only a raw-threshold replication.

The reverse-direction control uses the same recoverable images and degraded states. Under the image/text position partition, it is identical to restoring clean text states in the degraded run. The gap between its crossing and the repair limit is therefore a redundancy band, not an independent predictor. The independent gray-image crossing predicts held-out MMBench repair with 1.53-layer error, versus 3.65 for shared depth; excluding Gemma leaves 1.19 versus 2.30.

In the original five models, a linear probe and the logit lens remain readable after image-token restoration has become ineffective~\cite{huang2025causality}. Readability therefore does not establish causal repairability.

Our contributions are:
\begin{itemize}
    \item Repair under matched pixel degradation is sparse before it expires: 12--23 ranked encoder tokens reach half of full-patch recovery, while ten random orders need four to fifteen times as many.
    \item The repair window is model-specific and changes little across the tested matched corruptions. Gemma's benchmark shift marks the main exception.
    \item A clean-image gray-ablation crossing predicts held-out repair depth better than shared depth on seven models; the matched image/text control instead measures a model-specific redundancy band.
\end{itemize}

Restormer provides a non-oracle consistency check. It directly recovers 57\% of LLaVA's and 70\% of Qwen2-VL's oracle-recoverable noisy-image answers; its patched states lose effect after the oracle-defined repair limit. It removes the clean image as the patch source, not as the reference used to measure the boundary.

\section{Related Work}
\label{sec:related}

\paragraph{Patching and causal tracing.}
Causal tracing restores clean hidden states in a corrupted run~\cite{meng2022locating}. In VLMs, this intervention has localized grounded predictions in BLIP and traced visual and text states through decoder layers~\cite{10350956,li2026causal}; NOTICE replaces Gaussian embedding noise with structured corruptions~\cite{golovanevsky-etal-2025-vlms}. Recent failure analyses distinguish encoded information from causally used information through patching and source bounds~\cite{zhang2026encoded,veikherman2026encoded}. Intervention direction matters: clean-to-corrupt denoising tests sufficiency, whereas corrupt-to-clean noising tests necessity~\cite{heimersheim2024use}. Corruption and metric choices can also move the localized effect~\cite{zhang2024towards}. Under our full image/text partition, corrupt-to-clean image noising equals clean-to-corrupt text restoration; we interpret its offset from image restoration as a redundancy band.

\paragraph{Visual transfer and sparse tokens.}
Clean-image studies place image-to-text transfer in early or middle decoder layers~\cite{zhang2025cross,yin2025lifting,song2026does,Kaduri_2025_CVPR}; the path varies with model, task, and prompt~\cite{saporita2026fg,salazar2026pathways}. We compare the attention cutoff with necessity and repair rather than assume that all three crossings coincide; Gemma has a large MMBench point gap but no GQA separation. Other work finds that a small set of visual tokens stores or transfers much of the image information~\cite{neo2025towards,basu2024understanding}. FastV and VisionZip ask whether aggregate accuracy survives token removal~\cite{chen2024image,yang2025visionzip}; token pruning can also underperform random selection~\cite{wang2026token}. Register-like high-norm tokens offer another explanation for apparent sparsity~\cite{darcet2024vision}. We instead ask which clean states are sufficient to reverse a specific degraded answer, then compare gradient-based rankings with magnitude and high-norm controls.

\paragraph{Corruption and repair.}
Common-corruption benchmarks establish severity-controlled evaluation for vision models~\cite{hendrycks2019robustness}; MLLM-IC extends corruption testing to multimodal capabilities~\cite{qiu2025benchmarking}. Under adversarial corruption, decoder layers can amplify or suppress a corrupted visual signal~\cite{josephraj2026corrupted}. Pixel restoration methods such as Restormer provide a non-oracle source between the degraded and clean images~\cite{Zamir2021Restormer}. We test whether restored-image states preserve the oracle repair window. Readout experiments remain a control: decodable states need not control the answer, and feature alignment need not recover MLLM accuracy~\cite{huang2025causality,piorczynski2026limits}.

\section{Method}
\label{sec:method}

\begin{figure*}[t]
  \centering
  \includegraphics[width=\textwidth]{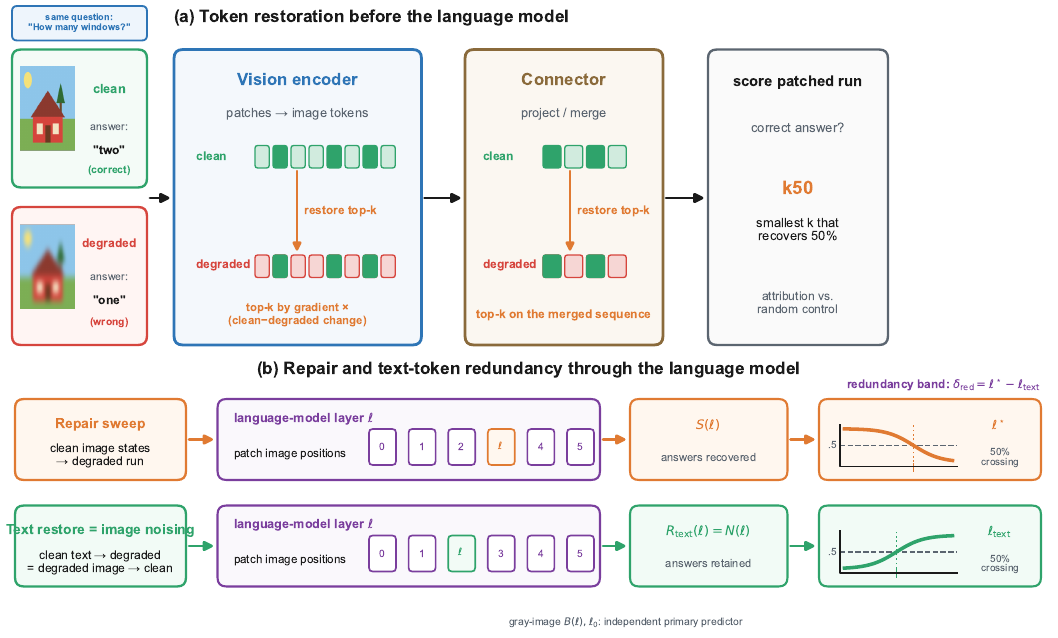}
  \caption{Causal repair measurements. (a) We select items answered correctly when clean and wrongly when degraded. Image tokens are ranked by the product of their clean--degraded change and the gradient toward the clean-image answer. The top-$k$ clean states replace degraded states; $k_{50}$ is the smallest $k$ reaching half of full-patch recovery. (b) Image restoration defines the repair crossing $\ell^\star$. Full image-position noising produces the same hybrid state as full text-position restoration, whose crossing is $\ell_{\mathrm{text}}$. Their difference $\delta_{\mathrm{red}}=\ell^\star-\ell_{\mathrm{text}}$ is the width of the redundancy band. The independent clean-image predictor remains the gray-ablation crossing $\ell_0$.}
  \label{fig:arch}
\end{figure*}

Figure~\ref{fig:arch} is the measurement. A degraded photo is passed through the vision encoder, the connector, and the language model. At each stage we put selected image tokens back to the values they took on the clean photo and test whether the answer returns. The clean photo is used only as the source of those values, and, in a separate pass, to predict where in the language model that restoration stops working.

\subsection{Paired runs and patching}
\label{sec:patching}

For an image $x$, question $q$, and answer $a$, a degradation $g$ gives $x_d=g(x)$. A forward pass on $x$ produces clean states $h^c_s$ at locus $s$; the same pass on $x_d$ produces degraded states $h^d_s$. We call an item recoverable when the clean run is correct and the degraded run is wrong. Option-shuffled copies are grouped by image; strict majority decides the image-level outcome, and a tie counts as failure.

For token positions $S$, the patch operator replaces only those degraded states:
\begin{equation}
  \widetilde h_{s,i}(S)=
  \begin{cases}
    h^c_{s,i}, & i\in S,\\
    h^d_{s,i}, & i\notin S.
  \end{cases}
  \label{eq:patch}
\end{equation}
The remaining layers run unchanged. A repair occurs only when the patched run returns the correct answer.

\subsection{Token ranking and $k_{50}$}
\label{sec:restore}

At locus $s$, let $h^c_{s,i}$ and $h^d_{s,i}$ be the clean and degraded states of image token $i$. We use the first-order attribution-patching approximation~\cite{syed-etal-2024-attribution} to rank tokens by their effect on the clean-image answer $a_c$:
\begin{equation}
  r_i =
  \left\langle
  h^c_{s,i}-h^d_{s,i},
  \nabla_{h^d_{s,i}}\log p(a_c\mid x_d,q)
  \right\rangle .
  \label{eq:rank}
\end{equation}
We replace the top-$k$ degraded states with their clean values for $k\in\{1,4,16,32,64,128,256\}$ and for the full token set. Let $R(k)$ be answer recovery at $k$ and $R_{\mathrm{full}}$ recovery from the full-token patch. The reported $k_{50}$ is the smallest $k$ for which $R(k)/R_{\mathrm{full}}\geq0.5$, linearly interpolated between adjacent tested values. Controls rank by clean--degraded norm, gradient norm, clean-state norm, and ten random orders; the control grid adds $k\in\{2,6,8,10,12,20,24\}$. A second attribution only pushes down the model's wrong degraded answer. Every repair is scored against the correct answer.

The main ranking does not use the ground-truth label: on a recoverable item, the model's clean-image answer $a_c$ is already correct. The ground-truth answer is used to select recoverable items and to score whether a patch repairs them.

The same ranking is applied in the vision encoder and after the connector. In the language model the token axis holds the whole prompt. We call the positions after the image block---the question, answer options, and answer cue---the \emph{text tokens}. We read image-token positions from the model's image-token marker, rather than assuming a fixed slice.

\subsection{The repair limit}
\label{sec:limit}

Inside the language model we sweep eight integer layers per model, with denser coverage in the early and middle layers. A patch at layer $\ell$ replaces the output of transformer block $\ell$, after its residual update; later blocks run normally. Image-token restoration replaces the complete image-token block with its clean state. Text-token restoration replaces every position after the image block. Let $R_{\mathrm{img}}(\ell)$ and $R_{\mathrm{text}}(\ell)$ be the shares of recoverable pairs whose answers return. With $R_{\mathrm{img}}(\mathrm{in})$ measured at the connector input, the normalized repair curve is
\begin{equation}
  S(\ell)=\frac{R_{\mathrm{img}}(\ell)}{R_{\mathrm{img}}(\mathrm{in})}.
\end{equation}
The text curve is a consistency check on the position partition, not an independent mediation result.

The repair limit is the first crossing
\begin{equation}
  \ell^\star = \min\{\ell : S(\ell)<0.5\},
  \label{eq:limit}
\end{equation}
linearly interpolated between adjacent swept layers. Decimal depths therefore denote estimated crossings, not physical layer indices.

On a clean image we run the opposite intervention. At layer $\ell$, the clean image-token states are replaced by states from a same-sized gray image. Let $\rho(\ell)$ be the share of clean answers that remain correct. Since some answers survive even when the image is removed at layer zero, we normalize by that empirical floor:
\begin{equation}
  B(\ell)=\frac{\rho(\ell)-\rho(0)}{1-\rho(0)} .
  \label{eq:retention}
\end{equation}
$B(\ell)$ is the share of image-dependent answers retained. The clean-image predictor is
\begin{equation}
  \ell_0 = \min\{\ell : B(\ell)\geq0.5\}.
  \label{eq:predictor}
\end{equation}
We define the lag as $\delta=\ell^\star-\ell_0$. To test a model not used to set this offset, we estimate the mean relative-depth lag on the other models and add it to that model's $\ell_0$.

\paragraph{Matched-contrast redundancy control.}
At layer $\ell$, split the residual stream into prefix $P$, image positions $I$, and later text positions $T$. The prefix precedes the image and is identical in clean and degraded runs. Patching degraded image states into the clean run gives $[P,I_d,T_c]$; patching clean text states into the degraded run gives the same $[P,I_d,T_c]$. Later blocks are unchanged, hence the two interventions are identical:
\begin{equation}
  N(\ell)\equiv R_{\mathrm{text}}(\ell).
\end{equation}
Their common crossing is $\ell_{\mathrm{text}}$. We normalize image restoration by its connector-input ceiling and text restoration by its connector-input floor, then define the redundancy-band width $\delta_{\mathrm{red}}=\ell^\star-\ell_{\mathrm{text}}$. It measures layers where either block alone suffices to recover the answer; it is not an independent predictor. The gray-image $\ell_0$ remains the primary clean-only predictor.

\subsection{Attention handoff check}
\label{sec:attention-check}

We compare repair with cumulative attention knockout on the same recoverable items in LLaVA, Qwen2-VL, and Gemma. In a clean run, the intervention blocks every text-query-to-image-key edge from layer $\ell$ onward. The cutoff is the first layer at which half of the floor-normalized clean answers survive. We use a predeclared $\pm2$-layer equivalence margin. Qwen3.5 is excluded because most of its blocks use recurrent Gated DeltaNet rather than full attention.

\paragraph{Non-oracle bridge.}
For Gaussian noise, Restormer's blind color-denoising checkpoint (\texttt{gaussian\_color\_denoising\_blind.pth}) supplies restored pixels without the clean image~\cite{Zamir2021Restormer}. We compare direct restored-pixel inference with restored-state and clean-oracle patches at the connector input, LM layers 8/12 for LLaVA, and layers 10/18 for Qwen2-VL. ``Oracle-recoverable'' denotes images answered correctly when clean, wrongly when noisy, and repaired by the clean connector-input patch.

\subsection{Readouts}
\label{sec:readouts}

At the same layers we train linear probes on the mean image-token state and the last prompt token. The target is the closed answer-label space of each MMBench family. A deterministic 30\% stratified test split is grouped by image, keeping option-shuffled copies together. We report degraded-to-clean accuracy ratios only when clean accuracy beats chance.

The logit lens applies the model's final normalization and output head to the mean and maximum image-token states and the last prompt token. It decodes family answer words from image states and the option letter from the last token. The final decoder layer must reproduce the model's output logits. We compare both readouts with $R_{\mathrm{img}}(\ell)$ at the same layers; readable states count as repair only when patching changes the answer.

\subsection{Uncertainty and prediction tests}
\label{sec:statistics}

Crossing intervals use 1,000 image-cluster bootstrap resamples. These intervals condition on the tested architectures. For the predictor advantage, we additionally resample models, compute a delete-one-model jackknife interval, and run an exact sign test over the seven per-model differences. We repeat crossing analysis at thresholds $0.25$, $0.5$, and $0.75$. Damage sensitivity regresses relative repair depth on accuracy drop with model fixed effects and a model-cluster bootstrap.

The primary prediction uses the independent gray-image crossing $\ell_0$ and is evaluated leave one model out. For held-out model $m$, the mean relative-depth lag is fitted on the other models and converted to layers using $m$'s layer count. The baseline assigns the mean relative repair depth and ignores $\ell_0$. MMBench-to-MMBench is primary; GQA is a replication, each over the same seven models. The exact test permutes model labels on the relative-depth gray crossings and refits all $7!$ leave-one-model-out assignments. Image-cluster intervals condition on this fixed architecture set; model jackknife and sign tests assess architecture sensitivity.

\subsection{Validity checks}
\label{sec:validity}

Every patch requires identical source and destination shapes and a hook that fires during the forward pass. Zeroing each configured locus must change at least one answer. Restoring the full clean tensor at the first patchable locus must reproduce the clean answer. For the logit lens, the final decoder layer must reproduce the model's top output token with cosine similarity above $0.99$. These checks prevent a dead module, an unfired hook, or a wrong normalization from appearing as a negative result.

\section{Experimental Setup}
\label{sec:experiments}

\subsection{Models}
\label{sec:models}

We use seven instruction-tuned models, listed in Table~\ref{tab:models}. Six have a vision encoder and a connector. Gemma 4 12B Unified has no encoder: a linear projection sends 280 image tokens directly into the language model. Its early-layer token counts motivate, but do not establish, the hypothesis that those layers absorb part of the encoder's role. The set spans 7B to 24B parameters and 28 to 48 language-model layers.

\begin{table}[t]
  \centering
  \caption{Models and image-token paths. Token counts are the image tokens that enter the language model. Gemma has no encoder.}
  \label{tab:models}
  \scriptsize
  \setlength{\tabcolsep}{3pt}
  \begin{tabular}{lccc}
    \hline
    Model & Vision path & Tokens & Layers \\
    \hline
    LLaVA-1.5-7B~\cite{liu2024improved} & CLIP & 576 & 32 \\
    Qwen2-VL-7B~\cite{wang2024qwen2} & $2\times2$ merger & 132 med. & 28 \\
    InternVL3.5-8B~\cite{wang2025internvl35advancingopensourcemultimodal} & one 448 px tile & 256 & 36 \\
    Qwen3.5-9B~\cite{qwen3.5} & $2\times2$ merger & 70 med. & 32 \\
    Gemma 4 12B~\cite{gemma4_2026} & none & 280 & 48 \\
    Idefics3-8B~\cite{laurençon2024building} & SigLIP, one view & 169 & 32 \\
    Mistral 3.1-24B~\cite{mistral_small_3_1_2025} & Pixtral + merger & 81 & 40 \\
    \hline
  \end{tabular}
\end{table}

We use one view or tile per image. LLaVA keeps all 576 CLIP patches; InternVL pixel-shuffles 1,024 patches to 256 tokens; Idefics3 and Mistral produce 169 and 81 tokens. Qwen2-VL and Qwen3.5 use dynamic resolution with the same pixel budget for each clean/degraded pair; Table~\ref{tab:models} gives their median merged counts. Qwen3.5 interleaves three recurrent Gated DeltaNet blocks with one full-attention block. Every patch has aligned source and destination shapes.

The eight swept language-model layers are LLaVA, Qwen3.5, and Idefics3: $\{0,2,4,6,8,12,16,24\}$; Qwen2-VL: $\{0,2,4,6,8,10,14,20\}$; InternVL3.5: $\{0,2,4,6,9,12,18,27\}$; Gemma 4: $\{0,3,6,9,12,18,24,36\}$; and Mistral-Small 3.1: $\{0,3,5,8,10,15,20,30\}$. Crossings between these integer layers are linearly interpolated; decimals do not denote physical layer indices.

The sparse grids define the primary analysis. A resolution check sweeps every layer in a local window covering both crossings for all seven MMBench models. Matched-contrast noising and denoising use the same recoverable images on MMBench and GQA. Attention knockout is run on LLaVA, Qwen2-VL, and Gemma; the latter two use dense local grids, and Gemma is tested on both benchmarks. The original ten-seed random control is at the encoder locus (and Gemma's first repairable LM locus). At the connector, we compare attribution with clean--degraded norm, gradient norm, clean-state norm, and ten random orders over all five degradations; Gemma again uses its first repairable LM locus.

The attention grids are $\{0,2,4,6,8,10,12,16,24\}$ for LLaVA, $\{0,7,\ldots,19\}$ for Qwen2-VL, and $\{0,17,\ldots,34\}$ for Gemma. The matched noising/text-restoration control uses each model's primary sparse grid; the resolution check uses the local dense windows listed in the supplement.

\subsection{Benchmarks and degradations}
\label{sec:data}

MMBench~\cite{liu2024mmbench} contributes 591 four-option questions over 149 images, divided into image emotion, nature relation, object localization, and social relation. From GQA~\cite{hudson2019gqa}, we use ChooseAttr, ChooseCat, and ChooseRel, with 300 two-option questions in each group. CV-Bench~\cite{tong2024cambrian} contains 2,638 questions: 788 counts, 650 spatial relations, 600 depth comparisons, and 600 distance comparisons.

The synthetic degradations are Gaussian blur, Gaussian noise, low light, JPEG compression, and downsampling, with severity levels 1--8. The MMBench stage map uses all five. The language-model layer sweeps use blur, noise, and low light on MMBench, GQA, and CV-Bench for the original five models. Idefics3 and Mistral-Small 3.1 run the same three degradations on MMBench and GQA. All stochastic degradations use seed 42.

ExDark~\cite{loh2019getting} provides a real low-light ablation, not a repair test. We use 543 images from the low (200), very-low (145), and twilight (198) bands across ten classes. With no paired clean photograph, ExDark cannot support clean-to-degraded token restoration. Across the original five models, its gray-state ablation crosses 2.25--2.57 layers earlier than the synthetic clean-image predictor, suggesting that real low light or the intervention shift moves the handoff earlier rather than supporting degradation invariance (supplement).

\subsection{Matched damage}
\label{sec:matched}

For each model, benchmark, and degradation, we first evaluate severities 5--8 on a deterministic, family-stratified sample of 200 questions. We select the lowest severity whose accuracy drop reaches 15 points; if none reaches the target, we retain the strongest tested severity and mark it as short.

The final filter is computed on the full evaluation set. A question family is kept only when its clean accuracy beats its option-set chance level in a one-sided $z$-test at $\alpha=0.01$. For the remaining families, let $\Delta$ be the paired clean-to-degraded accuracy drop and $\mathrm{SE}(\Delta)$ its standard error. We call a condition matched when
\begin{equation}
  \Delta + 1.645\,\mathrm{SE}(\Delta) \geq 0.15 .
  \label{eq:matched}
\end{equation}
The calibration produces 57 model-condition cells: 21 on MMBench, 21 on GQA, and 15 on CV-Bench. Of these, 47 pass the full-data matched rule: all 21 on MMBench, 17 on GQA, and 9 on CV-Bench. Matched-only figures and tables use those 47 cells. The descriptive damage-slope analysis uses the 42 normalized MMBench and GQA cells, including the four GQA cells marked short, and is labeled accordingly. LLaVA's CV-Bench distance family is removed because its clean accuracy equals chance. Drops above 30 points are retained but flagged as overshoots; Gemma low light on CV-Bench is the only such condition.

\subsection{Inference and scoring}
\label{sec:inference}

All models use deterministic greedy decoding and their released chat templates. We score the generated multiple-choice letter, not text copied from the prompt. Clean, degraded, and patched runs use the same question, option order, image processor, and numerical precision. LLaVA runs in FP16; the other models use BF16.

Option-shuffled copies of one image are grouped before recovery is counted. A clean/degraded or patched outcome is the majority result for that image, so repeating one image with different option orders does not create extra independent samples. When curves pool degradations, one image can contribute one image-condition unit per degradation; confidence intervals cluster those units by the underlying image. The experiments ran on A100 80GB GPUs. Each cell and each question family is saved independently, allowing interrupted sweeps to resume without recomputing completed results.

\section{Results}
\label{sec:results}

\subsection{A few ranked tokens repair the answer}
\label{sec:res-k50}

Table~\ref{tab:k50} reports the five-model stage-map experiment. In the four models with a vision encoder, restoring 12.2--22.6 attribution-ranked tokens recovers half of the answers recovered by a full patch. Ten random orders need roughly four to fifteen times as many tokens. The informed controls qualify this result: attribution has the lowest encoder $k_{50}$ for LLaVA and InternVL3.5, while gradient norm is lower for Qwen2-VL (11.1 versus 12.2) and Qwen3.5 (14.8 versus 22.6). Attribution is therefore consistently better than random at the encoder, but not uniformly optimal among informed rankings.

\begin{table}[t]
\centering
\caption{Token counts for half of full-patch recovery. Encoder attribution brackets are 95\% image-bootstrap intervals; connector attribution and informed controls use the denser review grid. Random is the median over ten orders. Dashes mark no 50\% crossing on the tested grid.}
\label{tab:k50}
\scriptsize
\textbf{Encoder locus (LLaVA L2C; others L2)}\\[-2pt]
\resizebox{\linewidth}{!}{%
\begin{tabular}{lrrrrr}
\hline
Model & Attr. & $\|\Delta h\|$ & $\|\nabla\|$ & High norm & Random \\
\hline
LLaVA-1.5 & 14.6 [12, 19] & 199.2 & 25.3 & -- & 126.5 \\
Qwen2-VL & 12.2 [11, 14] & -- & 11.1 & -- & 150.6 \\
InternVL3.5 & 13.4 [11, 16] & 99.4 & 45.2 & 95.0 & 199.9 \\
Qwen3.5 & 22.6 [18, 27] & 89.6 & 14.8 & -- & 89.4 \\
\hline
\end{tabular}}\\[2pt]
\textbf{Connector / early-LM locus (L3; Gemma L4)}\\[-2pt]
\resizebox{\linewidth}{!}{%
\begin{tabular}{lrrrrr}
\hline
Model & Attr. & $\|\Delta h\|$ & $\|\nabla\|$ & High norm & Random \\
\hline
LLaVA-1.5 & 11.2 & 208.6 & 28.9 & 252.2 & 126.5 \\
Qwen2-VL & 3.6 & 22.4 & 9.8 & 27.3 & 27.9 \\
InternVL3.5 & 3.5 & 18.5 & 9.2 & 26.8 & 29.4 \\
Qwen3.5 & 5.4 & 16.0 & 9.2 & 15.8 & 17.1 \\
Gemma 4 & 22.2 & 41.1 & 37.4 & 82.3 & 53.1 \\
\hline
\end{tabular}}
\end{table}

After the connector, attribution has the lowest $k_{50}$ in all five models. It needs 3.5--22.2 tokens, 1.7--2.7 times fewer than the strongest informed control and fewer than the median random order. The median Jaccard overlap at $k=16$ between attribution and high-norm tokens is 0.032--0.103 across models, compared with random expectations of 0.014--0.048; the sparse sets are not explained by high norm alone.

Gemma has no encoder. Its raw projection is a weak repair site (74.0 tokens versus 105.7 for the original random order), but at its first repairable language-model locus attribution needs 22.2 tokens versus 37.4 for gradient norm, 41.1 for clean--degraded norm, and 53.1 for random. This single-model pattern suggests that Gemma's first layers may take over part of the encoder's role. The small count is not caused by ranking toward the right answer: in a separate median-across-degradations analysis, the away-from-wrong ranking needs 13--26 encoder tokens, close to 11--23 for the clean-answer ranking.

\subsection{Repair stops as text-token recovery rises}
\label{sec:res-handoff}

\begin{figure*}[t]
  \centering
  \includegraphics[width=\textwidth]{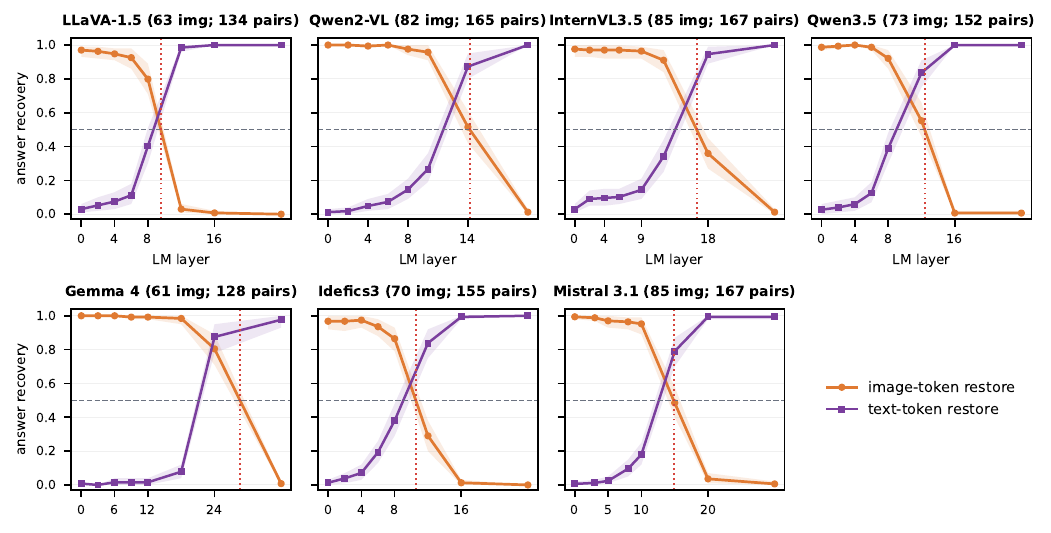}
  \caption{Layer-wise causal recovery on MMBench at matched damage (seven models). Titles report unique recoverable images and image-condition pairs. Orange restores image-token states. Purple restores text-token states and is algebraically identical to full image-position noising in the clean run. Curves show raw recovery; the dotted repair limit uses connector-input ceiling normalization. Shading is the 95\% image-cluster bootstrap interval.}
  \label{fig:handoff}
\end{figure*}

Figure~\ref{fig:handoff} shows the seven-model MMBench sweep after the matched-damage filter. Each model has 61--85 unique recoverable images, yielding 128--167 recoverable image-condition pairs across blur, noise, and low light. At the first swept layer, image-token restoration recovers 92--100\% of recoverable pairs, while text-token restoration recovers 1--3\%. The curves reverse with depth. At the last swept layer, image-token recovery is 0--2\% and text-token recovery is 99--100\%. The image curve defines the repair limit; the text curve checks the complementary position partition.

\subsection{Image and text blocks form a redundancy band}
\label{sec:res-attention}

Matched image-position noising is exactly the text-restoration intervention already shown in Fig.~\ref{fig:handoff}. Its gap from image restoration is the redundancy band where either image states or text states alone can recover the answer. On the dense grids, the mean widths range from 1.32 layers for LLaVA to 4.36 for Gemma (Table~\ref{tab:matched-contrast}).

\begin{table}[t]
\centering
\caption{Dense MMBench crossings on the same recoverable images. Image noising and text restoration are identical under the position partition; their crossing and the repair crossing bound the redundancy band.}
\label{tab:matched-contrast}
\scriptsize
\begin{tabular}{lrrr}
\hline
Model & Text crossing & Repair & Band \\
\hline
LLaVA-1.5 & 8.71 & 10.03 & 1.32 \\
Qwen2-VL & 12.33 & 14.09 & 1.76 \\
InternVL3.5 & 14.10 & 17.38 & 3.29 \\
Qwen3.5 & 9.57 & 12.95 & 3.38 \\
Gemma 4 & 22.04 & 26.39 & 4.36 \\
Idefics3 & 8.57 & 10.56 & 1.99 \\
Mistral 3.1 & 13.13 & 15.60 & 2.47 \\
\hline
\end{tabular}
\end{table}

The attention comparison is architecture- and task-dependent. LLaVA's mean attention cutoff is 10.09, near its repair crossing, with all three paired intervals inside the $\pm2$-layer margin. Qwen2-VL's cutoff is 12.94 versus dense repair at 14.09; the point gap lies inside the margin, but all three intervals cross it. Gemma MMBench has the only large point gap: 22.25 versus dense repair at 26.39. Two condition intervals establish separation and one crosses the margin, so the conservative benchmark decision is inconclusive. On GQA, Gemma's cutoff is 22.64 versus repair at 22.37; one condition is equivalent and two are inconclusive, with no evidence of separation.

\subsection{Restormer as a consistency check}
\label{sec:res-bridge}

Restormer directly corrects 56.5\% of LLaVA's oracle-recoverable noise errors (Wilson 95\% interval 42.2--69.8) and 70.1\% of Qwen2-VL's (58.3--79.8). These are 17.4\% and 31.5\% of all 149 images. A full patch at the connector input reproduces those same rates. At LM layers 8 and 10, before each repair limit, recovery is 39.1\% (26.4--53.5) and 68.7\% (56.8--78.5); after the limit it falls to 0.0\% and 9.0\%. This agrees with the oracle boundary but does not show that knowing the boundary improves restoration.

\subsection{Cross-condition variation and clean-image prediction}
\label{sec:res-predict}

Among conditions selected against the 15-point damage target, the normalized repair limit changes little across synthetic degradations. The largest within-model spread is 1.44 layers on MMBench and 1.10 on GQA. A within-model regression over these 42 normalized cells gives a slope of $0.0007$ relative depth per additional 0.1 accuracy drop; a model-cluster bootstrap gives a 95\% interval of $[-0.0115,0.0104]$. We detect no systematic shift inside this calibrated band; we did not sweep every severity through the language model.

The normalized MMBench/GQA comparison differs by at most 1.46 layers for six models. Gemma is the exception: its normalized sparse limits are 28.59 on MMBench and 22.37 on GQA. CV-Bench has only the legacy raw-threshold estimate, 22.80. The dense normalized MMBench estimate is 26.39, reducing the MMBench--GQA gap to 4.02 layers. Full intervals are in the supplement.

The independent gray-image crossing retains model-specific information. On MMBench, fitting its relative-depth lag on six models predicts the seventh with 1.53 layers of error, versus 3.65 for shared depth; exact permutation gives $p=0.005$. The 2.12-layer advantage has a model-resampling interval of $[1.03,3.51]$ and a delete-one-model jackknife interval of $[0.05,4.18]$. All seven per-model advantages are positive (two-sided sign test, $p=0.016$). Excluding Gemma leaves 1.19 versus 2.30. On GQA the errors are 1.53 versus 3.08. The text crossing gives 1.46 layers on MMBench, but this is internal consistency between the two curves in Fig.~\ref{fig:handoff}, not an independent prediction.

\begin{table}[t]
\centering
\caption{Leave-one-model-out localization on seven models. Gray-image blanking is the independent primary predictor. The text crossing comes from the same repair experiment and is descriptive only.}
\label{tab:prediction}
\scriptsize
\setlength{\tabcolsep}{3pt}
\begin{tabular}{lrrr}
\hline
Predictor & Ours & Shared depth & Exact $p$ \\
\hline
Gray-image blanking (primary) & 1.53 & 3.65 & 0.005 \\
Gray-image blanking, GQA & 1.53 & 3.08 & 0.016 \\
Text crossing (same experiment) & 1.46 & 3.65 & -- \\
\hline
\end{tabular}
\end{table}

\begin{figure*}[t]
  \centering
  \includegraphics[width=\textwidth]{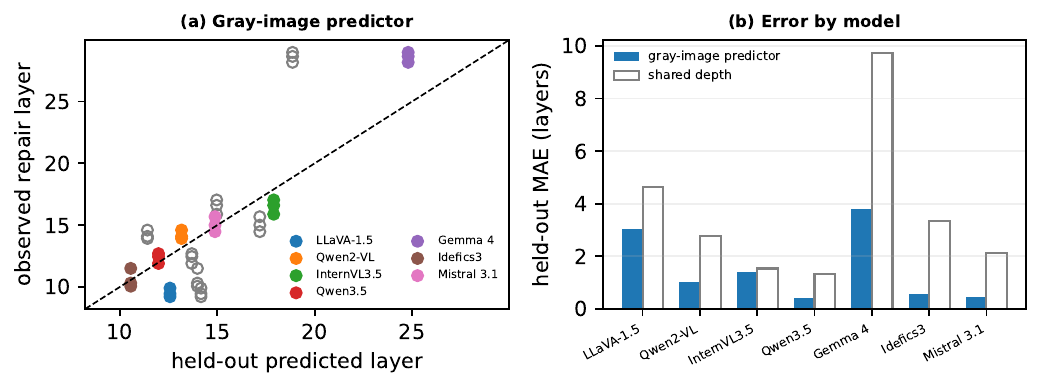}
  \caption{Held-out MMBench prediction from the independent gray-image crossing. Left: observed repair depth against the leave-one-model-out prediction; hollow gray markers show the shared-depth prediction for the same observations. Right: per-model error for the gray predictor and shared-depth baseline.}
  \label{fig:prediction}
\end{figure*}

\section{Ablations}
\label{sec:ablations}

Connector-ceiling normalization changes each MMBench crossing by at most 0.22 layers and changes predictor MAE from 1.55 to 1.53 (shared depth 3.66 to 3.65). Model resampling puts the normalized advantage in $[1.03,3.51]$ layers; delete-one-model jackknife gives $[0.05,4.18]$, and all seven per-model differences favor it ($p=0.016$). Without Gemma the MAEs are 1.19 and 2.30. The normalized condition-specific spread is at most 1.44 layers; raw-threshold population variants span 1.40--1.52. Dense-grid, threshold, nonmonotonicity, and ranking controls are in the supplement.

\section{Conclusion}
\label{sec:conclusion}

Under matched pixel degradation, a small ranked token set repairs answers before a model-specific expiry layer. Image noising and text restoration are the same intervention under the paper's position partition; their gap from image restoration measures a redundancy band. Gemma MMBench has the only large attention-to-repair point gap, while Gemma GQA does not; interval-based separation remains condition-dependent. The independent gray-image crossing predicts held-out repair depth better than shared depth. Restormer recovers more than half of oracle-recoverable noisy-image errors and provides a non-oracle consistency check on the boundary. These results cover decoder-only VLMs, synthetic paired degradations, multiple-choice answers, and fixed prompts; open-ended generation and naturally paired photographs remain untested.

\section{LLM Usage Statement}
We used Cursor and its integrated GPT-5.6 Sol model to assist with research ideation, literature organization, drafting and language editing, simulated critical review, and the implementation and debugging of experiment, analysis, testing, and figure-generation code. All numerical results in this paper come from executed code and saved experiment outputs; no generative model was used to synthesize experimental data. The authors reviewed the generated code, prose, figures, and analyses, and manually reviewed the bibliography and citation keys. The authors take full responsibility for the accuracy, reproducibility, and integrity of the submitted work.

{
  \small
  \bibliographystyle{ieeenat_fullname}
  \bibliography{main}
}

\clearpage
\appendix
\renewcommand{\thesection}{\Alph{section}}
\renewcommand{\thesubsection}{\thesection.\arabic{subsection}}
\setcounter{section}{0}
\maketitlesupplementary
\section{Overview}

This supplement reports the resolution, population, baseline, and ranking checks that support the main paper. Section~\ref{sec:supp-crossing} examines how the repair crossing changes with interpolation, threshold, benchmark, and architecture. Sections~\ref{sec:supp-attention}--\ref{sec:supp-real} give the attention, population, damage-strength, and real-dark-image checks. Sections~\ref{sec:supp-readouts}--\ref{sec:supp-sparse} report the readout, restored-image, and sparse-token controls.

\section{Crossing and Prediction Robustness}
\label{sec:supp-crossing}

\subsection{Crossing resolution}

The primary analysis linearly interpolates the ceiling-normalized 50\% crossing between adjacent swept layers. This gives 1.53 layers of MMBench leave-one-model-out error for the gray-image predictor, compared with 3.65 for shared relative depth. Connector normalization changes any individual MMBench crossing by at most 0.22 layers. On the raw-threshold curves, isotonic interpolation gives 1.55 versus 3.66, while selecting the first measured layer below threshold gives 3.44 versus 4.64.

Sub-grid decimals are estimates rather than physical layer indices. Treating each crossing as its measured layer interval gives 0.27 layers of mean distance from the prediction to the observed interval; using interval midpoints gives 2.80 layers. Table~\ref{tab:supp-influence} reports the held-out prediction and measured interval for every model.

\begin{table}[t]
\centering
\caption{Primary MMBench leave-one-model-out prediction. Pred. and Obs. mean average the three conditions; MAE averages the three degradation-specific absolute errors. Brackets are the union of measured crossing intervals.}
\label{tab:supp-influence}
\scriptsize
\setlength{\tabcolsep}{2pt}
\begin{tabular}{lrrrr}
\hline
Held model & Pred. & Obs. mean & MAE & Measured bin \\
\hline
LLaVA-1.5 & 12.6 & 9.5 & 3.0 & [8,12] \\
Qwen2-VL & 13.2 & 14.2 & 1.0 & [10,20] \\
InternVL3.5 & 17.9 & 16.5 & 1.4 & [12,18] \\
Qwen3.5 & 12.0 & 12.3 & 0.4 & [8,16] \\
Gemma 4 & 24.8 & 28.6 & 3.8 & [24,36] \\
Idefics3 & 10.6 & 10.6 & 0.6 & [8,12] \\
Mistral 3.1 & 14.9 & 15.0 & 0.5 & [10,20] \\
\hline
\end{tabular}
\end{table}

\subsection{Benchmark and threshold sensitivity}

Table~\ref{tab:stability} reports the repair crossing by benchmark. Six models differ by at most 1.46 layers between normalized MMBench and GQA estimates. Gemma is the exception: its dense normalized MMBench estimate is 26.39, compared with 22.37 on GQA. CV-Bench has only the raw-threshold replication.

\begin{table}[t]
\centering
\caption{Mean interpolated repair-limit layer by benchmark. MMBench and GQA use connector-ceiling normalization; CV-Bench is the legacy raw-threshold replication. Brackets span per-condition 95\% intervals.}
\label{tab:stability}
\scriptsize
\resizebox{\linewidth}{!}{%
\begin{tabular}{lcccc}
\hline
Model & MMB & GQA & CV-B & Max gap \\
\hline
LLaVA-1.5 & 9.5 [8.6,10.2] & 9.0 [8.3,9.3] & 9.1 [8.6,9.3] & 0.5 \\
Qwen2-VL & 14.2 [13.2,15.7] & 14.3 [13.2,15.4] & 13.6 [13.2,14.3] & 0.7 \\
InternVL3.5 & 16.5 [15.1,18.7] & 16.5 [15.8,17.3] & 16.4 [15.8,16.9] & 0.2 \\
Qwen3.5 & 12.3 [10.9,13.4] & 11.6 [10.9,12.5] & 11.3 [10.9,11.8] & 1.1 \\
Gemma 4 & 28.6 [26.4,29.8] & 22.4 [21.6,23.0] & 22.8 [22.1,23.5] & 6.2 \\
Idefics3 & 10.6 [9.6,12.5] & 9.9 [9.0,10.9] & -- & 0.7 \\
Mistral 3.1 & 15.0 [13.6,16.8] & 13.6 [12.8,14.8] & -- & 1.4 \\
\hline
\end{tabular}}
\end{table}

Changing the crossing threshold does not reverse the MMBench result, but the GQA predictor is worse than shared depth at threshold 0.25 (Table~\ref{tab:thresholds}). We therefore treat the 0.5 crossing as the predeclared primary estimand rather than claim threshold-independent prediction.

\begin{table}[t]
\centering
\caption{Threshold sensitivity of the primary gray-image predictor. Values are leave-one-model-out MAE in layers (predictor versus shared depth).}
\label{tab:thresholds}
\scriptsize
\begin{tabular}{lrr}
\hline
Threshold & MMBench & GQA \\
\hline
0.25 & 1.84 vs. 4.46 & 4.01 vs. 3.86 \\
0.5 & 1.53 vs. 3.65 & 1.53 vs. 3.08 \\
0.75 & 1.26 vs. 3.30 & 1.16 vs. 3.07 \\
\hline
\end{tabular}
\end{table}

\subsection{Alternative estimators and baselines}

Table~\ref{tab:ablations} collects interpolation, population, damage-slope, and ranking checks. Table~\ref{tab:supp-boundaries} reports the gray-image crossing, repair limit, and lag for every model. The lag is model dependent; the predictor fits its mean relative-depth offset only on the six training models in each leave-one-model-out fold.

\begin{table}[t]
\centering
\caption{Checks on crossing resolution, population choice, and token ranking. The linear row uses the normalized primary seven-model MMBench pairing; isotonic and first-sampled rows are legacy raw-threshold checks. Population rows report the largest within-model degradation spread. The ten-seed random row reports the range of model-level pooled medians.}
\label{tab:ablations}
\scriptsize
\setlength{\tabcolsep}{1.5pt}
\resizebox{\linewidth}{!}{%
\begin{tabular}{lll}
\hline
Check & Setting & Outcome \\
\hline
Crossing & normalized linear & 1.53 vs. 3.65 MAE \\
Crossing & isotonic & 1.55 vs. 3.66 MAE \\
Crossing & first sampled & 3.44 vs. 4.64 MAE \\
Crossing & raw dense LLaVA+Gemma & 1.14 vs. 3.22 MAE \\
Population & condition-specific & 1.44 max spread \\
Population & common images & 1.52 max spread \\
Population & equal family & 1.40 max spread \\
Damage slope & normalized model-cluster boot. & [-0.0115,0.0104] per 0.1 drop \\
Ranking & away-from-wrong & 13--26 encoder tokens \\
Ranking & clean-answer & 11--23 encoder tokens \\
Ranking & random, 10 seeds & 89--200 encoder tokens \\
\hline
\end{tabular}}
\end{table}

\begin{table}[t]
\centering
\caption{Primary sparse-grid MMBench crossings by model. $\ell_0$ is the clean-image crossing, $\ell^\star$ is the mean degraded repair crossing, and $\delta=\ell^\star-\ell_0$.}
\label{tab:supp-boundaries}
\scriptsize
\begin{tabular}{lrrr}
\hline
Model & $\ell_0$ & $\ell^\star$ & $\delta$ \\
\hline
LLaVA-1.5 & 9.25 & 9.52 & 0.27 \\
Qwen2-VL & 10.78 & 14.20 & 3.42 \\
InternVL3.5 & 14.44 & 16.51 & 2.07 \\
Qwen3.5 & 9.14 & 12.35 & 3.21 \\
Gemma 4 & 21.00 & 28.59 & 7.59 \\
Idefics3 & 7.68 & 10.62 & 2.94 \\
Mistral 3.1 & 11.29 & 15.01 & 3.72 \\
\hline
\end{tabular}
\end{table}

Single-feature architecture baselines do not explain the result. Layer count, parameter count, image-token count, and connector type give 3.70--7.20 layers of error (Table~\ref{tab:supp-architecture}). Holding out Qwen2-VL and Qwen3.5 together gives 0.86 layers for the clean-crossing predictor and 1.75 for shared depth. Seven models are insufficient for a reliable multivariable architecture model.

\begin{table}[t]
\centering
\caption{Primary MMBench leave-one-model-out baselines. Each architecture baseline fits one feature at a time on the other six models.}
\label{tab:supp-architecture}
\scriptsize
\begin{tabular}{lr}
\hline
Predictor & MAE (layers) \\
\hline
Clean crossing + lag & 1.53 \\
Clean crossing, zero lag & 3.32 \\
Shared relative depth & 3.65 \\
LM layer count & 4.43 \\
Parameter count & 7.20 \\
Image-token count & 5.46 \\
Connector type & 3.70 \\
\hline
\end{tabular}
\end{table}

The image bootstrap conditions on the tested architectures. Table~\ref{tab:model-uncertainty} therefore adds model resampling, a delete-one-model jackknife, an exact sign test, and the no-Gemma comparison.

\begin{table}[t]
\centering
\caption{Architecture sensitivity of the primary gray-image predictor on MMBench. Advantage is shared-depth MAE minus predictor MAE.}
\label{tab:model-uncertainty}
\scriptsize
\begin{tabular}{lr}
\hline
Check & Result \\
\hline
All seven models & 2.12-layer advantage \\
Model bootstrap & [1.03,3.51] \\
Delete-one-model jackknife & [0.05,4.18] \\
Per-model sign test & 7/7, $p=0.016$ \\
Without Gemma & 1.10-layer advantage \\
\hline
\end{tabular}
\end{table}

\subsection{Association is not identity}

The gray-image and repair crossings use different interventions and image populations. A diagnostic recomputes both on each condition's recoverable-image set and divides degraded recovery by its first-layer value. Across 42 model--benchmark--condition cells, the crossings correlate at $r=0.95$, but their lag ranges from $-1.65$ to $7.60$ layers. This supports association, not identity.

\begin{table}[t]
\centering
\caption{Layer-0 values. $R_{\rm img}(0)$ is degraded full-block recovery; $\rho(0)$ is clean-answer retention after inserting gray-image states.}
\label{tab:supp-layer0}
\scriptsize
\begin{tabular}{lrrrr}
\hline
& \multicolumn{2}{c}{MMBench} & \multicolumn{2}{c}{GQA} \\
Model & $R_{\rm img}(0)$ & $\rho(0)$ & $R_{\rm img}(0)$ & $\rho(0)$ \\
\hline
LLaVA-1.5 & 0.97 & 0.36 & 0.91 & 0.56 \\
Qwen2-VL & 1.00 & 0.28 & 1.00 & 0.51 \\
InternVL3.5 & 0.98 & 0.30 & 0.99 & 0.45 \\
Qwen3.5 & 0.99 & 0.29 & 0.98 & 0.81 \\
Gemma 4 & 1.00 & 0.03 & 0.99 & 0.04 \\
Idefics3 & 0.97 & 0.20 & 0.95 & 0.60 \\
Mistral 3.1 & 0.99 & 0.16 & 0.98 & 0.46 \\
\hline
\end{tabular}
\end{table}

Dense matched sweeps cover all seven MMBench models. Mean repair crossings range from 10.03 layers for LLaVA to 26.39 for Gemma; population-level image/text band widths range from 1.32 to 4.36 layers. The complete dense values appear in Table~3 of the main paper.

\begin{table}[t]
\centering
\caption{Targeted resolution and handoff checks. Dense rows compare the sparse-grid MMBench mean with the normalized dense-grid mean. Attention rows compare LLaVA's cumulative attention-knockout crossing with its dense repair crossing on the same recoverable images.}
\label{tab:supp-priority-checks}
\scriptsize
\setlength{\tabcolsep}{3pt}
\begin{tabular}{llrrr}
\hline
Check & Condition & Reference & New & $|\Delta|$ \\
\hline
Dense grid & LLaVA-1.5 & 9.52 & 10.03 & 0.51 \\
Dense grid & Gemma 4 & 28.59 & 26.39 & 2.20 \\
Attention cutoff & gaussian blur & 9.75 & 10.35 & 0.60 \\
Attention cutoff & gaussian noise & 10.11 & 9.33 & 0.77 \\
Attention cutoff & low light & 10.24 & 8.70 & 1.54 \\
\hline
\end{tabular}
\end{table}

\section{Attention-Handoff Checks}
\label{sec:supp-attention}

\begin{table}[t]
\centering
\caption{Same-item attention cutoff versus repair crossing. Equivalence requires the paired gap interval to lie inside $\pm2$ layers.}
\label{tab:attention-handoff}
\scriptsize
\resizebox{\linewidth}{!}{%
\begin{tabular}{llrrrl}
\hline
Model & Benchmark & Attention & Repair & Gap & Decision \\
\hline
Gemma 4 & GQA & 22.64 & 22.37 & 0.27 & inconclusive \\
Gemma 4 & MMBENCH & 22.25 & 26.39 & -4.14 & inconclusive \\
LLaVA-1.5 & MMBENCH & 10.09 & 10.03 & 0.06 & equivalent \\
Qwen2-VL & MMBENCH & 12.94 & 14.09 & -1.16 & inconclusive \\
\hline
\end{tabular}}
\end{table}

The attention comparison resamples paired images and uses a predeclared $\pm2$-layer equivalence margin. LLaVA is equivalent in all three MMBench conditions. Qwen2-VL's point gaps lie inside the margin, but its intervals cross it. Gemma MMBench has a 4.14-layer mean point gap: two condition intervals establish separation and one crosses the margin. On GQA, one condition establishes equivalence and two remain inconclusive.

\begin{figure*}[t]
  \centering
  \includegraphics[width=\textwidth]{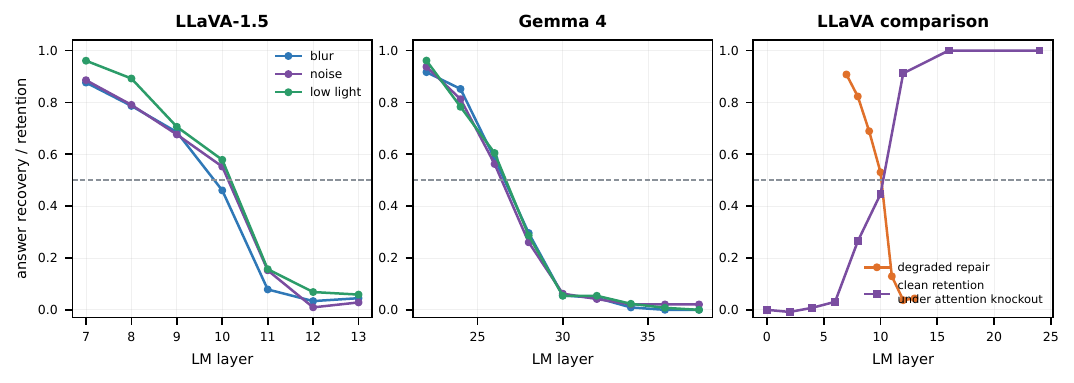}
  \caption{Targeted layer-resolution and attention checks. Left and center: dense image-token repair curves for LLaVA and Gemma under the three language-model degradations. Right: condition-averaged LLaVA degraded-image repair and clean-image retention under cumulative text-query-to-image-key attention knockout.}
  \label{fig:supp-priority}
\end{figure*}

\section{Population and Damage Checks}
\label{sec:supp-population}

\subsection{Recoverable-population alignment}

The main analysis defines recoverability separately for each degradation. A raw-threshold sensitivity check instead uses the intersection of images recoverable under blur, noise, and low light. Common sets contain 25--106 images per model and benchmark. The largest within-model spread is 1.52 layers on the common population, compared with 1.44 in the normalized primary analysis; equal family weights give 1.40. Item composition therefore does not explain the small within-benchmark spread.

\subsection{Damage-strength dependence}

The primary slope uses the 42 connector-normalized MMBench/GQA cells. It is $0.0007$ relative depth per additional 0.1 accuracy drop, with a seven-model cluster-bootstrap interval of $[-0.0115,0.0104]$. The analysis only covers the calibrated band and does not establish invariance to arbitrary damage.

\section{Real Dark Photographs}
\label{sec:supp-real}

ExDark has no paired clean photograph, so it cannot support the clean-to-degraded restoration intervention. We instead replace image-token states in the dark-image run with states from a gray image and measure when the answer ceases to depend on those states. This ablation crossing occurs earlier than the synthetic clean-image predictor: by 2.57 layers using the MMBench predictor and 2.25 using GQA, averaged over the five stage-map models. The offset is similar across low, very-low, and twilight bands. We report ExDark separately because it changes the task and intervention as well as the image distribution.

\section{Readout Controls}
\label{sec:supp-readouts}

We train linear probes on the mean image-token state and the last prompt token at the swept layers. Targets are the closed answer-label spaces of the MMBench families. A deterministic 30\% stratified test split is grouped by image, keeping option-shuffled copies together. Ratios are reported only when clean accuracy exceeds chance.

The logit lens applies the model's final normalization and output head to mean and maximum image-token states and to the last prompt token. It decodes answer words from image states and the option letter from the final prompt token. The final decoder layer reproduces the model's output logits. Table~\ref{tab:readouts} shows that readability can remain after image-token patches stop repairing the answer.

\begin{table}[t]
\centering
\caption{Readability at causal depths (five models). Values are percentages. Probe values use the deepest swept layer; lens values are the range of median post-limit retention across three readouts.}
\label{tab:readouts}
\scriptsize
\setlength{\tabcolsep}{2.5pt}
\begin{tabular}{lrrrr}
\hline
Model & Restore & Probe-img & Probe-last & Lens range \\
\hline
LLaVA-1.5 & 0.0 & 79 & 83 & 85--94 \\
Qwen2-VL & 1.2 & 86 & 86 & 80--86 \\
InternVL3.5 & 1.2 & 88 & 86 & 73--98 \\
Qwen3.5 & 0.7 & 89 & 83 & 72--88 \\
Gemma 4 & 0.8 & 83 & 78 & 67--81 \\
\hline
\end{tabular}
\end{table}

\section{Restored-Image Consistency Check}
\label{sec:supp-restored}

For Gaussian noise, Restormer's blind color-denoising checkpoint supplies restored pixels without the clean image. We compare direct restored-pixel inference with restored-state and clean-oracle patches at the connector input, at LM layers 8/12 for LLaVA, and at layers 10/18 for Qwen2-VL. ``Oracle-recoverable'' images are correct when clean, wrong when noisy, and repaired by the clean connector-input patch.

\begin{table}[t]
\centering
\caption{Restormer consistency check on Gaussian noise. Brackets are Wilson 95\% intervals on oracle-recoverable images; All is the share of all 149 images repaired.}
\label{tab:restored-bridge}
\scriptsize
\resizebox{\linewidth}{!}{%
\begin{tabular}{lrrrrr}
\hline
Model & Restored pixels & All & Connector & Before & After \\
\hline
LLaVA-1.5 & 56.5 [42.2,69.8] & 17.4 & 56.5 [42.2,69.8] & 39.1 & 0.0 \\
Qwen2-VL & 70.1 [58.3,79.8] & 31.5 & 70.1 [58.3,79.8] & 68.7 & 9.0 \\
\hline
\end{tabular}}
\end{table}

Direct Restormer inference corrects 56.5\% of LLaVA's and 70.1\% of Qwen2-VL's oracle-recoverable errors. Patching the complete restored state at the connector reproduces direct inference. The restored-state patch loses effect after the oracle-defined repair limit, which is a consistency check rather than evidence that knowing the limit improves restoration.

\section{Sparse-Repair Controls}
\label{sec:supp-sparse}

The main paper compares attribution with clean--degraded norm, gradient norm, clean-state norm, and ten random orders at the encoder and connector loci over five degradations. Informed controls use the denser $k$ grid specified in Method. A dash in Table~2 of the main paper means that pooled recovery did not cross 50\% on the tested grid.

At the connector or early-LM locus, attribution has the lowest $k_{50}$ in all five models. It needs 1.7--2.7 times fewer tokens than the strongest informed control. At the encoder, attribution is lowest for LLaVA and InternVL3.5, while gradient norm is lower for Qwen2-VL and Qwen3.5. The result therefore supports sparse repair and consistent improvement over random ordering, not universal optimality among informed rankings. The median Jaccard overlap at $k=16$ between attribution and high-norm tokens is 0.032--0.103 across models, compared with random expectations of 0.014--0.048.

\begin{table}[t]
\centering
\caption{Connector-level $k_{50}$ ranking controls, pooled over five degradations. Random is the median over ten orders.}
\label{tab:ranking-controls}
\scriptsize
\resizebox{\linewidth}{!}{%
\begin{tabular}{lrrrrr}
\hline
Model & Attribution & $\|\Delta h\|$ & $\|\nabla\|$ & Clean norm & Random \\
\hline
Gemma 4 & 22.2 & 41.1 & 37.4 & 82.3 & 53.1 \\
InternVL3.5 & 3.5 & 18.5 & 9.2 & 26.8 & 29.4 \\
LLaVA-1.5 & 11.2 & 208.6 & 28.9 & 252.2 & 126.5 \\
Qwen2-VL & 3.6 & 22.4 & 9.8 & 27.3 & 27.9 \\
Qwen3.5 & 5.4 & 16.0 & 9.2 & 15.8 & 17.1 \\
\hline
\end{tabular}}
\end{table}

The recoverable-item filter changes the denominator. Each encoder model has 745 evaluated image-condition units over five degradations. Full restoration repairs 13.7--19.7\% of all units. Restoring the top 16 attribution-ranked tokens repairs 3.6--8.1\% of all units, or 23.9--46.9\% of recoverable units. Across the original 320 MMBench cell-ranking curves, 25 contain an adjacent decrease, so $k_{50}$ is a linearly interpolated sufficient-set summary rather than a monotone cardinality law.

\begin{figure*}[t]
  \centering
  \includegraphics[width=\textwidth]{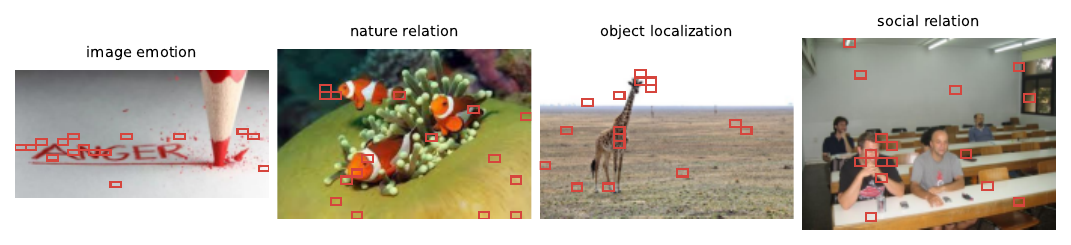}
  \caption{Four Gaussian-blur examples from the LLaVA connector control. Red boxes mark the 16 highest-ranked attribution tokens. These are descriptive examples, not evidence that attribution consistently selects object boxes.}
  \label{fig:supp-token-maps}
\end{figure*}

\begin{figure*}[t]
  \centering
  \includegraphics[width=\textwidth]{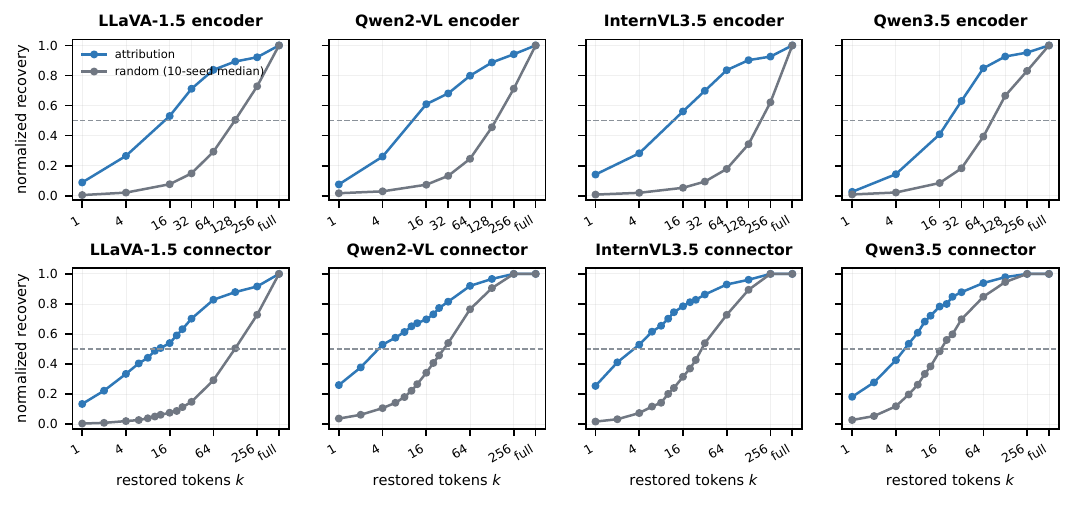}
  \caption{Matched-severity sparse-repair curves on the common tested $k$ grid. Lines pool MMBench cells, and the random curve is the median over ten orders. Tick labels are actual token counts; the final point restores the full token set.}
  \label{fig:supp-k50}
\end{figure*}

\end{document}